%% file: templateArxiv.tex
\documentclass{article}

\usepackage{PRIMEarxiv}

\input{math_commands.tex} %All the command are in this file

\usepackage{amsmath}
\usepackage{mathtools}
\usepackage[utf8]{inputenc} % allow utf-8 input
\usepackage[T1]{fontenc}    % use 8-bit T1 fonts
\usepackage{natbib}         % for \citet and \citep commands
\usepackage{hyperref}       % hyperlinks
\usepackage{url}            % simple URL typesetting
\usepackage{booktabs}       % professional-quality tables
\usepackage{amsfonts}       % blackboard math symbols
\usepackage{nicefrac}       % compact symbols for 1/2, etc.
\usepackage{comment}        % enables the comment environment
\usepackage{microtype}      % microtypography
\usepackage{lipsum}
\usepackage{fancyhdr}       % header
\usepackage{graphicx}       % graphics
\graphicspath{{media/}}     % organize your images and other figures under media/ folder
\usepackage{tikz}           % for TikZ figures

\usepackage{mathrsfs}
\usepackage{hyperref}
\usepackage{url}
\usepackage{amsmath}
\usepackage{algorithm}
\usepackage{algorithmic}
\usepackage{subfig}
\usepackage{amssymb}

\newcommand{\rhotr}{\rho^{\rm tr}}
\newcommand{\rhote}{\rho^{\rm te}}
\title{
Importance Weighting Correction of Regularized Least-Squares for Target Shift
}

\author{
  Davit Gogolashvili\\
  Weierstrass Institute for
Applied Analysis and Stochastics \\
  Berlin, Germany\\
  \texttt{davit.gogolashvili@wias-berlin.de} 
}

\begin{document}
\maketitle

\begin{abstract}
    Importance weighting is a standard tool for correcting distribution shift, but its statistical 
    behavior under {\em target shift}—where the label distribution changes between training and testing 
    while the conditional distribution of inputs given the label remains stable—remains under-explored. We analyze importance-weighted 
    kernel ridge regression under target shift and show that, because the weights depend only on 
    the output variable, reweighting corrects the train–test mismatch without altering the input-space
    complexity that governs kernel generalization. Under standard RKHS regularity and capacity 
    conditions and a mild Bernstein-type moment condition on the label weights, we obtain 
    finite-sample guarantees showing that the estimator achieves the same convergence behavior as in 
    the no-shift case, with shift severity affecting only the constants through weight moments. We 
    complement these results with matching minimax lower bounds, establishing rate optimality and 
    quantifying the unavoidable dependence on shift severity. We further study more general weighting schemes
    and prove that weight misspecification induces an {\em irreducible bias}: 
    the estimator concentrates around an induced population regression function that generally 
    differs from the desired test regression function unless the weights are accurate. Finally, 
    we derive consequences for plug-in classification under target shift via standard calibration 
    arguments.
\end{abstract}

% keywords can be removed

\section{Introduction}\label{sec:intro}

Modern learning systems are routinely deployed in environments where the data distribution evolves between training and deployment. This mismatch---often grouped under \emph{dataset shift}---appears in domain adaptation, off-policy evaluation, active learning, and many scientific workflows where training data are collected under one regime and predictions are required under another. When the training and test distributions differ, standard empirical risk minimization targets the wrong population risk and can yield systematically biased predictions.

A principled and widely used correction is \emph{importance weighting} (IW): training examples are reweighted 
by the likelihood ratio between test and training distributions so that the weighted training risk matches the 
desired test risk. While importance weighting under \emph{covariate shift} (changing input marginal with invariant 
conditional labels) has been extensively analyzed for kernel methods and beyond, the corresponding 
theory for \emph{target shift}, where the label marginal changes but the conditional input distribution given the label remains invariant,
is substantially less complete at the level of sharp rates and minimax optimality for nonparametric regression.

The goal of the paper is to understand, for kernel ridge regression under target shift:
(i) whether IW can achieve the same statistical behavior as in the no-shift case under mild tail assumptions on $w_Y(Y)$;
(ii) whether these guarantees are rate-optimal in a minimax sense, including their dependence on shift severity; and
(iii) what happens when the weights are not exact, as is unavoidable in practice.

We provide a finite-sample and minimax analysis of importance-weighted kernel ridge regression (IW-KRR) under target shift.

\begin{enumerate}
\item \textbf{Finite-sample guarantees under mild weight moments.}
Under standard RKHS assumptions (a source condition controlling regression regularity and an effective-dimension 
condition controlling capacity) and a Bernstein-type moment condition on the label weights, we prove high-probability 
bounds for IW-KRR in test $L^2(\rhote_X)$ error. The resulting convergence matches the classical behavior of kernel 
ridge regression in the no-shift case; shift affects only the constants through weight-moment parameters.

\item \textbf{Minimax optimality with explicit shift severity.}
We complement the upper bounds with matching minimax lower bounds over natural target-shift classes. In particular, 
when the label weights are uniformly bounded by a parameter $W$, the lower bound quantifies the unavoidable dependence 
on $W$, showing that the upper bound scaling in shift severity is not an artifact of analysis, but fundamental limitation 
of learning under target shift.

\item \textbf{Incorrect or estimated weights induce an irreducible bias.}
We analyze weighted KRR with a generic weighting function in place of the true target-shift ratio. We show that misspecification changes the \emph{population target}: the estimator concentrates around an induced regression function that generally differs from the desired test regression function. This yields an explicit, non-vanishing bias term in test $L^2(\rhote_X)$ error that disappears only when the weights are correct.

\item \textbf{Consequences for classification.}
For binary labels, we translate the regression guarantees into plug-in classification bounds via standard calibration and margin arguments, yielding fast rates under Tsybakov-type noise conditions.
\end{enumerate}

Our analysis follows an operator-theoretic approach to kernel ridge regression. The key observation is the classical 
unbiasedness identity: importance weighting transforms expectations under the training distribution into 
expectations under the test distribution. As a result, the weighted empirical covariance and 
cross-covariance operators concentrate around the \emph{test} operators. This leads to sharp bounds driven by the 
effective dimension of the test covariance operator and by moments of the label weights. For incorrect weights, 
we develop a refined excess-risk decomposition that cleanly separates (a) the stochastic error terms controlled 
by operator concentration from (b) a population-level mismatch term reflecting the difference between the induced 
and desired regression functions.

\paragraph{Related work.}
The idea of correcting dataset shift by reweighting training examples has a long history: one rewrites the target
(test) risk as a weighted expectation under the source (training) distribution, and then studies how the resulting
variance inflation depends on the weight behavior. For \emph{covariate shift} and sample-selection bias this viewpoint
goes back to early work such as \citet{shimodaira2000}, and it has since developed into a broad literature on density
ratio estimation and reweighted empirical risk minimization, including nonparametric analysis for kernel methods and
regularized least squares (e.g., \citet{cortes2010, ma2022optimally, gogolashvili2023importance}). In parallel,
a large body of work focuses on how to stabilize reweighting when weights are large or heavy-tailed, through
regularization, clipping, or robustification, and on understanding how reweighting interacts with model
complexity and the geometry induced by the input distribution.

Target shift has been studied most extensively in the \emph{classification} setup.
Early approaches adjust posteriors to new class priors via EM-type procedures \citep{saerens2002adjusting}, while more recent
methods estimate the shifted label proportions from unlabeled target data using a trained predictor, often through
moment matching or confusion-matrix inversion \citep{lipton2018detecting} and its regularized variants
\citep{azizzadenesheli2019regularized}. Subsequent work provides unified perspectives and finite-sample analyses of
label-shift estimators \citep{garg2020unified}, and highlights the importance of calibration and likelihood-based
procedures for strong practical performance \citep{alexandari2020maximum}. Minimax theory for label shift has also
been developed in nonparametric classification, clarifying fundamental limits when labels are missing at test time
\citep{maity2020minimax}; closely related, the quantification literature studies estimating class prevalences under
prior shift and includes lower-bound style guarantees \citep{vaz2019quantification}.

Compared to classification, there is less work addressing target shift with \emph{continuous} labels.
Importance-weight estimation under target shift for continuous targets 
has been addressed via kernel mean matching~\citep{zhang2013domain} and  $L^2$-distance-based distribution 
matching~\citep{nguyen2016continuous}, both avoiding explicit density  estimation. While these works focus on estimating the weights, our 
results complement them by providing a finite-sample and minimax 
analysis of importance-weighted kernel ridge regression under target 
shift. We also characterize the effect of arbitrary (possibly 
misspecified) weights: these induce a different population regression 
target, leading to an explicit bias term that persists unless the 
weights are accurate. As a further consequence, we derive plug-in 
classification rates under standard calibration and margin conditions.

\paragraph{Organization.}
Section~\ref{sec:prelims} introduces the RKHS framework and assumptions. Section~\ref{sec:target_shift} presents the main target-shift bounds and the minimax lower bound. Section~\ref{sec:incorrect_weights} studies incorrect weights and derives the irreducible bias phenomenon. Section~\ref{sec:binary_classification} gives the classification corollaries, and Section~\ref{sec:simulations} illustrates the theory empirically.

\section{Problem Setup and Preliminaries}\label{sec:prelims}

We study supervised learning when the data distribution changes between training and deployment. Concretely, we observe
i.i.d.\ training samples $(x_1,y_1),\dots,(x_n,y_n)$ drawn from a training distribution $\rhotr$ on $Z=X\times Y$, where 
$X$ is a measurable input space, and $Y \subset \mathbb{R}$ is the output space.
We want our predictor to perform well under a (different) test distribution $\rhote$. Our target is the
\emph{test regression function},
\[
f_{\rhote}(x):= \int y d\rhote(y\mid x),
\]
and we evaluate error in the test $L^2(\rhote_X)$ norm, equivalently the excess squared-loss risk under $\rhote$.

Throughout the paper, we consider two types of dataset shifts:
\begin{itemize}
    \item \textbf{Covariate shift}: The conditional output distribution given the input is 
    the same in training and test domains, while the input marginal distribution changes:
    \[
    \rhotr(x,y)=\rho(y\mid x)\rhotr_X(x),
    \,\,
    \rhote(x,y)=\rho(y\mid x)\rhote_X(x).
    \]

    \item \textbf{Target shift}: The conditional input distribution given the 
    label is the same in training and test domains, while the label marginal distribution changes:
    \[
    \rhotr(x,y)=\rho(x\mid y)\rhotr_Y(y),
    \,\,
    \rhote(x,y)=\rho(x\mid y)\rhote_Y(y).
    \]
\end{itemize}

Under absolute continuity $d\rhote\ll d\rhotr$, we define importance weight (or Radon--Nikodym derivative) as
\[
w(x,y)=\frac{d\rhote}{d\rhotr}(x,y).
\]
Under covariate shift this weight depends only on $x$, whereas under target shift it depends only on $y$.

We work in an RKHS $\mathcal H$ over $X$ with reproducing kernel $K$ satisfying $\sup_{x\in X}K(x,x)\le \kappa\le 1$.
Given a positive weighting function $w:Z\to\mathbb R_+$ and $\lambda>0$, importance weighted kernel ridge regression (IW-KRR) is
\begin{equation}\label{iw_emp_risk_clean}
f_{\mathbf{z},\lambda}^{\rm IW}
:=\argmin_{f\in\mathcal H}\left\{\frac1n\sum_{i=1}^n w(x_i,y_i)\big(f(x_i)-y_i\big)^2+\lambda\|f\|_{\mathcal H}^2\right\}.
\end{equation}
The solution exists, it is  unique and has the form (see \citet{smale2004shannon})
\begin{equation} \label{ERM_IW}
    f_{\mathbf{z},\lambda}^{\rm IW}=\left(S_{\mathbf{x}}^{T}  M_{\mathbf{w}} S_{\mathbf{x}}+\lambda \right)^{-1} S_{\mathbf{x}}^{T}  M_{\mathbf{w}}\mathbf{y},
\end{equation}
where $M_{\mathbf{w}}$ is the diagonal matrix with entries $w(x_i,y_i)$, the sampling operator $S_{\mathbf x}:\mathcal H\to\mathbb R^n$,
$(S_{\mathbf x}f)_i:=f(x_i)$, and its adjoint $S_{\mathbf x}^\top a=\frac{1}{n}\sum_{i=1}^n a_i K_{x_i}$, where $K_x:=K(x,\cdot)$.

We define the covariance operator $T_{\nu}:\mathcal{H} \rightarrow \mathcal{H}$ and integral operator $L_{\nu}:L^2(\nu) \rightarrow \mathcal{H}$ 
for measure $\nu$ as
\begin{align*}
    \left(T_{\nu} f\right)(x) &= \int K(x', x) f(x')\, d\nu(x'), \quad f \in \mathcal{H},\\
    \left(L_{\nu} f\right)(x) &= \int K(x', x) f(x')\, d\nu(x'), \quad f \in L^2(\nu).
\end{align*}
When $\nu=\rhote_X$, we write $T$ and $L$ for brevity. Under boundedness, $T_{\nu}$ is positive trace class with $\|T_{\nu}\| \leq 1$. 

Replacing the empirical risk in \eqref{iw_emp_risk_clean} by its expectation and using $w = d\rhote/d\rhotr$, the population problem reduces to
$$
f_\lambda = \argmin_{f \in \mathcal{H}} \left\{ \|f - f_{\rhote}\|_{\rhote_X}^2 + \lambda \|f\|_{\mathcal{H}}^2 \right\},
$$
whose unique solution is
\begin{equation}\label{eq:pop_solution}
f_\lambda = (T + \lambda)^{-1} T f_{\mathcal{H}},
\end{equation}
where $f_{\mathcal{H}}$ is the projection of the regression function $f_{\rhote}$ onto
the closure of $\mathcal{H}$ in $L^2(\rhote)$:
\[
f_{\mathcal{H}} = \argmin_{f \in \mathcal{H}} \|f - f_{\rhote}\|^2_{\rhote_X}.
\]

Throughout this paper, we assume the importance weighting function is known. When unknown, it can 
be estimated from labeled training data and unlabeled test samples using target shift adaptation methods 
for continuous responses \citep{zhang2013domain, nguyen2016continuous}.

We adopt standard assumptions from the kernel learning literature characterizing 
the regularity of the target function and the marginal distribution:

\begin{assumption}[Source Condition]\label{source_condition}
There exist $r \in [1/2,1]$ and $R>0$ such that $\left\|L^{-r} f_{\mathcal{H}}\right\|_{\rhote_X} \leq R.$ 
\end{assumption}

The source condition quantifies the regularity of the target function $f_{\mathcal{H}}$ 
relative to the integral operator $L = L_{\rhote_X}$. Writing 
$f_{\mathcal{H}} = \sum_{j \geq 1} \langle f_{\mathcal{H}}, e_j \rangle_{\rhote_X} e_j$ 
in the eigenbasis $(e_j)_{j \geq 1}$ of $L$, the condition 
$\|L^{-r} f_{\mathcal{H}}\|_{\rhote_X} \leq R$ is equivalent to
\begin{equation*}
  \sum_{j \geq 1} \mu_j^{-2r} \langle f_{\mathcal{H}}, e_j \rangle_{\rhote_X}^2 \leq R^2,
\end{equation*}
which constrains the Fourier coefficients of $f_{\mathcal{H}}$ to decay at a rate 
controlled by the eigenvalues $(\mu_j)_{j \geq 1}$ of $L$. Larger values of $r$ impose 
faster decay of these coefficients, meaning that $f_{\mathcal{H}}$ is increasingly 
concentrated on the leading eigendirections of $L$ and is thus effectively smoother 
relative to the geometry induced by the kernel and the marginal $\rhote_X$.

Equivalently, Assumption~\ref{source_condition} can be written as 
$f_{\mathcal{H}} \in \mathrm{Range}(L^r)$, i.e., there exists 
$g \in L^2(\rhote_X)$ with $\|g\|_{\rhote_X} \leq R$ such that 
$f_{\mathcal{H}} = L^r g$. This range condition is a standard device in inverse problems 
and regularization theory \citep{engl1996regularization}, where $r$ plays the role of a 
smoothness index relative to the forward operator. In the learning-theoretic context, it 
was introduced by \citet{smale2007learning} and has since become a standard regularity 
assumption in the analysis of kernel methods 
\citep{caponnetto2007optimal, steinwart2009optimal}.

The parameter $r$ admits a natural interpretation through the interpolation scale of the 
RKHS. The case $r = 1/2$ corresponds to $f_{\mathcal{H}} \in \mathcal{H}$, the standard 
well-specified setting where the target function lies in the RKHS itself. Values 
$r > 1/2$ encode additional smoothness beyond membership in $\mathcal{H}$: for instance, 
when $\mathcal{H}$ is a Sobolev space $H^{\alpha}(\mathbb{R}^d)$, the source condition 
with parameter $r$ corresponds to $f_{\mathcal{H}} \in H^{2r\alpha}(\mathbb{R}^d)$ under 
the uniform measure on $[0,1]^d$, provided $2r\alpha > d/2$.

\begin{assumption}[Effective Dimension]\label{ass:eff_dim}
For some $s \in (0,1]$,
\begin{equation*}
  E_s:=1 \vee \sup _{\lambda \in(0,1]} \sqrt{\mathcal{N}(\lambda) \lambda^{s}} < \infty,
\end{equation*}
where $\mathcal{N}(\lambda)=\operatorname{Tr}\left[T(T+\lambda)^{-1} \right]$ is the effective dimension \citep{caponnetto2007optimal}.
\end{assumption}

The effective dimension acts as a soft count of the eigenvalues of 
$T = T_{\rhote_X}$ that exceed the threshold $\lambda$. Writing 
$\mathcal{N}(\lambda) = \sum_{j \geq 1} \mu_j / (\mu_j + \lambda)$ in terms of the 
eigenvalues $(\mu_j)_{j \geq 1}$ of $T$, each summand transitions smoothly from 
$\approx 1$ when $\mu_j \gg \lambda$ to $\approx 0$ when $\mu_j \ll \lambda$. 
Assumption~\ref{ass:eff_dim} requires that 
$\mathcal{N}(\lambda) \lesssim \lambda^{-s}$, controlling how fast this effective 
complexity grows as the regularization level $\lambda \to 0$.

This condition is intimately linked to the spectral decay of $T$. For polynomial eigenvalue 
decay $\mu_j \asymp j^{-b}$, one has $\mathcal{N}(\lambda) \asymp \lambda^{-1/b}$, so 
Assumption~\ref{ass:eff_dim} holds with $s = 1/b$. Smaller values of $s$ thus correspond 
to faster eigenvalue decay and, intuitively, to a lower intrinsic complexity of the RKHS 
relative to the marginal $\rhote_X$. For Sobolev spaces $H^{\alpha}(\mathbb{R}^d)$ with 
the uniform measure on $[0,1]^d$, the eigenvalues decay as $\mu_j \asymp j^{-2\alpha/d}$, 
giving $s = d/(2\alpha)$, which reflects the familiar interplay between smoothness and 
dimensionality.

The boundary case $s=1$ is distinguished: the bound 
$\mathcal{N}(\lambda) \lambda \leq \operatorname{Tr}[T] \leq \kappa^2$ holds universally 
under bounded kernels, so Assumption~\ref{ass:eff_dim} with $s=1$ imposes no restriction on 
the spectral decay. This yields capacity-independent results in the sense of 
\citet{smale2007learning}, at the cost of slower convergence rates. In contrast, when the 
eigenvalue decay is known to be sufficiently fast, taking $s < 1$ leads to improved rates 
that reflect the favorable geometry of the problem.

We emphasize that the effective dimension is defined with respect to the \emph{test} marginal 
$\rhote_X$, not the training marginal $\rhotr_X$. In the dataset shift setting, this 
distinction is essential: the spectral properties of $T_{\rhote_X}$ and $T_{\rhotr_X}$ may 
differ substantially depending on how different the underlying distributions are. 
Our convergence rates therefore depend on the complexity of the target domain as seen through 
the lens of the RKHS, which is the natural quantity governing the difficulty of prediction 
under the test distribution.

% ===========================================================
%  SECTION 2 — Target Shift Analysis
% ===========================================================

\section{Main Results: Target Shift Analysis}\label{sec:target_shift}

We now present a finite-sample analysis of IW-KRR under \emph{target shift} and establish minimax-optimal rates.
A key structural feature is that, under target shift,
\[
w(x,y)=\frac{d\rhote(x,y)}{d\rhotr(x,y)}=\frac{d\rhote_Y(y)}{d\rhotr_Y(y)}=:w_Y(y),
\]
so importance weights act only on the \emph{output} variable. As a consequence, reweighting corrects the mismatch
between $\rhotr$ and $\rhote$ without altering the input-space geometry encoded by the covariance operator $T$ and
its effective dimension.

Our convergence  analysis relies on a Bernstein-type conditional moment bound for $w_Y(Y)$.

\begin{assumption}[Target-shift weight moments]\label{ass:target_shift_weights}
Let $w_Y=d\rhote_Y/d\rhotr_Y$. There exist positive constants $W_Y$ and $\sigma_Y$ such that for all integers $m\ge 2$,
\begin{equation*}
\sup_{x\in X}\ \int w_Y^{\,m-1}(y)\,d\rhote_Y(y\mid x)
\;\le\; \frac{1}{2}\,m!\,W_Y^{m-2}\sigma_Y^{2}.
\end{equation*}
\end{assumption}

\begin{remark}
Assumption~\ref{ass:target_shift_weights} is a standard Bernstein-moment condition (uniformly over $x$) and is
satisfied, for example, when $w_Y(Y)$ is uniformly bounded or sub-exponential under $\rhote(\cdot\mid X=x)$.
\end{remark}
\begin{remark}
    Unlike covariate shift where boundedness conditions involve the parameter $q \in [0,1]$ controlling 
    tail behavior (see \citet[Assumption~3]{gogolashvili2023importance}), target shift requires only a 
    fixed moment condition. This structural difference underlies the superior robustness of target shift 
    that we establish below.
\end{remark}

\paragraph{Why IW targets the test operators under target shift.}
The following unbiasedness identity is the basic reason IW correction works particularly cleanly 
in the target-shift
setting: for any measurable $g:X\to\mathbb{R}$,
\begin{align*}
    \int w_Y(y)\,g(x) d\rhotr(x,y)
&=\iint w_Y(y)\,g(x)\,d\rho(x\mid y)\,d\rhotr_Y(y)\ \\
&=\iint g(x)d\rho(x\mid y)\,d\rhote_Y(y)\\
&=\int g(x)\,d\rhote_X(x).
\end{align*}
Applying this with $g(x)=K_x$ (vector-valued) and $g(x)=K_x\otimes K_x$ (operator-valued) shows that the weighted
empirical quantities in \eqref{ERM_IW} converge to the \emph{test} operators:
\[
S_{\mathbf{x}}^{T}  M_{\mathbf{w}}\mathbf{y} \rightarrow \int yw_Y(y)\,K_x d\rhotr(x,y)=L f_{\rhote},
\]
and 
\[
S_{\mathbf{x}}^{T}  M_{\mathbf{w}} S_{\mathbf{x}} \rightarrow \int w_Y(y)\,(K_x\otimes K_x) d\rhotr(x,y)=T.
\]
Thus $f^{\rm IW}_{\mathbf{z},\lambda}$ concentrates around the same regularized solution \eqref{eq:pop_solution} as in the no-shift case.

\begin{theorem}[IW-KRR under Target Shift]\label{IW_KRR_TarS}
Let $\rhote$ and $\rhotr$ be distributions on $X\times[-M,M]$ satisfying target shift and
Assumptions~\ref{source_condition}, \ref{ass:eff_dim}, \ref{ass:target_shift_weights}.
Assume $\lambda\le \|T\|$ and set, for $\delta\in(0,1)$,
\begin{equation}\label{opt_reg_ts}
\lambda = \left(\frac{8E_s(\sqrt{W_Y}+\sigma_Y)\log\!\left(\frac{6}{\delta}\right)}{\sqrt{n}}\right)^{\frac{2}{2r+s}}.
\end{equation}
Then, with probability at least $1-\delta$,
\begin{equation}\label{generalization_bound_ts}
\|f^{\rm IW}_{\mathbf{z},\lambda}-f_{\mathcal{H}}\|_{\rhote_X}
\leq C \left(\frac{8E_s(\sqrt{W_Y}+\sigma_Y)\log\!\left(\frac{6}{\delta}\right)}{\sqrt{n}}\right)^{\frac{2r}{2r+s}},
\end{equation}
where $C = 3(M+R)$.
\end{theorem}

See Appendix~\ref{proof:target_shift} for the proof. Theorem~\ref{IW_KRR_TarS} provides a high-probability bound of the test $L^2(\rhote_X)$ error for IW-KRR under
target shift. The bound decomposes the difficulty of the problem into (i) the intrinsic approximation/regularity
governed by the source condition parameter $r$ and the eigen-decay parameter $s$ (via $E_s$), and (ii) the
distributional mismatch across domains, which enters only through the output-space weight moments
$(W_Y,\sigma_Y)$. Remarkably, the convergence rates $\mathcal{O}(n^{-\frac{r}{2r+s}})$ are the same as in the classical (no-shift) kernel 
regression analysis \citep{caponnetto2007optimal}, and the effect of shift appears through the constants $(W_Y,\sigma_Y),$ both in the rates 
and in the optimal regularization level \eqref{opt_reg_ts}.

If the weight ratio is uniformly bounded, $w_Y(y)\le W$, then 
Assumption~\ref{ass:target_shift_weights} holds with 
$W_Y \leq W$ and $\sigma_Y \leq W$, and \eqref{generalization_bound_ts} yields the explicit scaling
\[
\|f^{\rm IW}_{\mathbf{z},\lambda}-f_{\mathcal{H}}\|_{\rhote_X}
\lesssim
\Bigl(\frac{W}{n}\Bigr)^{\frac{r}{2r+s}}.
\]
In this sense, $W$ acts as an effective sample-size penalty through the concentration of weighted empirical
operators, while the smoothness-driven exponent remains unchanged.

Importantly, the dependence on $W$ in the above rate is not an artifact of the proof technique: it is
\emph{minimax-optimal} under target shift. In particular, even when the regression function satisfies the same
source condition and the kernel has the same effective-dimension behavior, no estimator can in general improve the
scaling in $W$ uniformly over the class of target-shift pairs with $w_Y\le W$. The next result (proved in Appendix~\ref{proof:lower_bound_my_proof}) 
provides a matching minimax lower bound of order $(W/n)^{\frac{r}{2r+s}}$.

\begin{theorem}[Minimax Lower Bound for Target Shift]\label{theo:lower_bound_ts} 
    There exists a pair of distributions $(\rhotr,\rhote)$ satisfying target shift and Assumptions~\ref{source_condition} and~\ref{ass:eff_dim}, 
    such that, for any estimator $\hat{f}$ based on $n$ training samples, we have
\begin{equation}\label{lower_bound_rate}
\inf_{\hat{f}} \sup_{f_{\mathcal{H}}\in\mathcal{F}_r(R)}
\mathbb{E}\!\left[\|\hat{f}-f_{\mathcal{H}}\|_{\rhote_X}\right]
\geq C\,
\left(\frac{W}{n}\right)^{\frac{r}{2r+s}},
\end{equation}
where $\mathcal{F}_r(R)=\{f\in\mathcal{H}:\|L^{-r}f\|_{\rhote_X}\le R\},$ $r \geq 1/2$ and $C>0$ depends only on $(M,R,s,r)$.
\end{theorem}

Together, Theorems~\ref{IW_KRR_TarS} and~\ref{theo:lower_bound_ts} establish that IW-KRR is minimax-optimal under target shift over the class $\mathcal{F}_r(R)$.

It is instructive to compare the target-shift rates with those obtained under covariate shift. 
Under covariate shift, the importance weights $w(x) = d\rhote_X/d\rhotr_X$ act on the 
\emph{input} variable and directly interact with the kernel geometry: they reshape the 
covariance operator and inflate the effective dimension, leading to degraded convergence 
rates when the weights are heavy-tailed \citep{gogolashvili2023importance}. In contrast, 
under target shift the weights depend only on the output $y$ and leave the covariance 
operator $T$ and the effective dimension $\mathcal{N}(\lambda)$ unchanged. The distributional 
mismatch therefore enters the bound \eqref{generalization_bound_ts} solely through the 
scalar constants $(W_Y, \sigma_Y)$, preserving the no-shift exponent 
$n^{-\frac{r}{2r+s}}$. Notably, when the weights are uniformly bounded by $W$, both 
covariate and target shift yield the same minimax rate $\mathcal{O}((W/n)^{\frac{r}{2r+s}})$, so the 
dependence on the shift severity parameter $W$ is identical in both settings. The key 
difference lies in \emph{how} this dependence arises: under covariate shift, $W$ interacts 
with the input-space complexity and can cause more severe deterioration when the weights 
have heavy tails, whereas under target shift, $W$ enters only as a scalar prefactor, 
leaving the smoothness--complexity tradeoff intact.

% ===========================================================
%  SECTION 4 — INCORRECT WEIGHTS: BIAS ANALYSIS
% ===========================================================

\section{Incorrect Weights: Bias Analysis}\label{sec:incorrect_weights}

In the previous section, we established learning guarantees under target shift assuming access to the \emph{correct} importance weights.  
We now turn to a more fundamental and practically relevant issue: in practice, these weights are typically estimated and therefore potentially approximate.  
What are the statistical consequences of using a weighting function that deviates from the true target-shift ratio \(w_Y = d\rhote_Y / d\rhotr_Y\)?

Throughout this section we consider weighting schemes of the form 
$$v_Y(y) = \frac{d\rho'_Y(y)}{d\rhotr_Y(y)},$$ 
where $\rho'$ is an
arbitrary probability measure on $Y$ absolutely continuous w.r.t. $ \rhotr_Y $. Denote by $f^{v}_{\mathbf z,\lambda}$ the W-KRR estimator
\eqref{ERM_IW} with weights $w(x,y)=v_Y(y)$. Our goal is to quantify the test error
$\|f^{v}_{\mathbf z,\lambda}-f_{\mathcal H}\|_{\rhote_X}$ under target shift.

Before giving the precise generalization bound, we first characterize the population-level target induced by the 
misspecified weights $v_Y$. This is given by the following proposition, which 
shows that using incorrect weights $v_Y\neq w_Y$ induces a regression function that differs from the true regression function $f_{\rhote}.$

\begin{comment}
\begin{proposition}\label{prop:induced_target}
Assume target shift and let $w_Y := d\rhote_Y / d\rhotr_Y$.
For any measurable $v_Y : Y \to (0,\infty)$, define the ratio 
$\eta(y) := v_Y(y)/w_Y(y)$ and the functions
\begin{equation}\label{induced_regression_components}
\phi(x) := \int y\,\eta(y)\,d\rhote(y \mid x), \,\,
\psi(x) := \int \eta(y)\,d\rhote(y \mid x).
\end{equation}
Then the population weighted risk 
$\mathcal{R}_v(f) := \int v_Y(y)(f(x)-y)^2\,d\rhotr(x,y)$ 
admits the representation
\begin{equation}\label{eq:Rv_as_test}
\mathcal{R}_v(f) = \int \eta(y)(f(x)-y)^2\,d\rhote(x,y),
\end{equation}
and its minimizer is the induced regression function
\begin{equation}\label{eq:induced_regression_short}
f^\eta(x) = \frac{\phi(x)}{\psi(x)}.
\end{equation}
In particular, if $v_Y = w_Y$ then $\eta \equiv 1$ and $f^\eta = f_{\rhote}$.
\end{proposition}

\begin{proof}
Under target shift, for any integrable $h$,
\begin{align*}
    \int v_Y(y)\,h(x,y)\,d\rhotr(x,y)
&= \int w_Y(y)\,\eta(y)\,h(x,y)\,d\rhotr(x,y) \\
&= \int \eta(y)\,h(x,y)\,d\rhote(x,y).
\end{align*}

Taking $h(x,y) = (f(x)-y)^2$ gives \eqref{eq:Rv_as_test}.
To identify the minimizer, condition on $x$: since $\psi(x) > 0$, the measure 
$d\rho^\eta(y \mid x) := \eta(y)\,d\rhote(y \mid x)/\psi(x)$ is a probability distribution, and
\begin{multline*}
\int \eta(y)(f(x)-y)^2\,d\rhote(y \mid x) \\
= \psi(x)\int (f(x)-y)^2\,d\rho^\eta(y \mid x).
\end{multline*}
The right-hand side is minimized pointwise by the conditional mean 
$\int y\,d\rho^\eta(y \mid x) = \phi(x)/\psi(x)$.
\end{proof}
\end{comment}
\begin{proposition}\label{prop:induced_target}
Assume target shift and let $w_Y := d\rhote_Y / d\rhotr_Y$.
For any measurable $v_Y : Y \to (0,\infty)$, define the ratio
$\eta(y) := v_Y(y)/w_Y(y)$ and the functions
\begin{equation}\label{induced_regression_components}
\phi(x) := \int y\,\eta(y)\,d\rhote(y \mid x), \qquad
\psi(x) := \int \eta(y)\,d\rhote(y \mid x).
\end{equation}
Then the population weighted risk
$\mathcal{R}_v(f) := \int v_Y(y)(f(x)-y)^2\,d\rhotr(x,y)$
admits the representation
\begin{equation}\label{eq:Rv_as_test}
\mathcal{R}_v(f) = \int \eta(y)(f(x)-y)^2\,d\rhote(x,y),
\end{equation}
and its (unconstrained) minimizer is the induced regression function
\begin{equation}\label{eq:induced_regression_short}
f^\eta(x) = \frac{\phi(x)}{\psi(x)}.
\end{equation}
Moreover, define the \emph{population $\mathcal H$-minimizer} (equivalently, the $\psi$-weighted projection of $f^\eta$)
by
\begin{equation}\label{eq:fHv_def_in_prop}
f_{\mathcal H}^{\eta}\in \argmin_{f\in\mathcal H}\mathcal R_v(f)
\;=\;
\argmin_{f\in\mathcal H}\big\|\sqrt{\psi}\,(f-f^\eta)\big\|_{L_2(\rhote_X)}^{2}.
\end{equation}
It satisfies the normal equation
\begin{equation}\label{eq:fHv_normal_equation_in_prop}
LM_\psi f_{\mathcal H}^{\eta}=L\phi.
\end{equation}
\end{proposition}

\begin{proof}
Under target shift, for any integrable $h$,
\begin{align*}
    \int v_Y(y)\,h(x,y)\,d\rhotr(x,y)
&= \int w_Y(y)\,\eta(y)\,h(x,y)\,d\rhotr(x,y) \\
&= \int \eta(y)\,h(x,y)\,d\rhote(x,y).
\end{align*}
Taking $h(x,y) = (f(x)-y)^2$ gives \eqref{eq:Rv_as_test}.

To identify the (unconstrained) minimizer, condition on $x$: since $\psi(x) > 0$, the measure
$d\rho^\eta(y \mid x) := \eta(y)\,d\rhote(y \mid x)/\psi(x)$ is a probability distribution, and
\begin{equation*}
\int \eta(y)(f(x)-y)^2\,d\rhote(y \mid x)
= \psi(x)\int (f(x)-y)^2\,d\rho^\eta(y \mid x).
\end{equation*}
The right-hand side is minimized pointwise by the conditional mean
$\int y\,d\rho^\eta(y \mid x) = \phi(x)/\psi(x)$, yielding \eqref{eq:induced_regression_short}.

For the $\mathcal H$-restricted minimizer, use the conditional variance decomposition:
for every $x$,
\[
\int \eta(y)(f(x)-y)^2\,d\rhote(y\mid x)
=
\psi(x)\bigl(f(x)-f^\eta(x)\bigr)^2
+
\psi(x)\int \bigl(y-f^\eta(x)\bigr)^2\,d\rho^\eta(y\mid x),
\]
where the second term does not depend on $f$. Integrating over $x$ shows that
$\mathcal R_v(f)$ differs from $\|\sqrt{\psi}\,(f-f^\eta)\|_{L_2(\rhote_X)}^{2}$ by an additive constant,
hence \eqref{eq:fHv_def_in_prop}.

Finally, to obtain the normal equation \eqref{eq:fHv_normal_equation_in_prop}, let $g\in\mathcal H$ and
differentiate the map $t\mapsto \mathcal R_v(f_{\mathcal H}^v+t g)$ at $t=0$ using \eqref{eq:Rv_as_test}:
\begin{align*}
0
&=\left.\frac{d}{dt}\right|_{t=0}\int \eta(y)\bigl(f_{\mathcal H}^v(x)+t g(x)-y\bigr)^2\,d\rhote(x,y) \\
&=2\int \eta(y)\bigl(f_{\mathcal H}^v(x)-y\bigr)g(x)\,d\rhote(x,y) \\
&=2\int \bigl(\psi(x)f_{\mathcal H}^v(x)-\phi(x)\bigr)g(x)\,d\rhote_X(x) \\
&=2\big\langle LM_\psi f_{\mathcal H}^v - L\phi,\; g\big\rangle_{\mathcal H},
\end{align*}
which implies $LM_\psi f_{\mathcal H}^v = L\phi$.
\end{proof}

\begin{comment}
\begin{proposition}\label{prop:induced_target}
Assume target shift and let $w_Y:=d\rhote_Y/d\rhotr_Y$.
Fix any measurable $v_Y:Y\to(0,\infty)$ and define $\eta(y):=v_Y(y)/w_Y(y)$. The minimizer of the population weighted risk 
$$\mathcal R_v(f):=E_{\rhotr}[v_Y(Y)(f(X)-Y)^2]$$ 
is the induced regression function
\begin{equation}\label{eq:induced_regression_short}
f^\eta(x)
=\frac{E_{\rhote}[Y\eta(Y)\mid X=x]}{E_{\rhote}[\eta(Y)\mid X=x]}
=:\frac{\phi(x)}{\psi(x)},
\end{equation}

In particular, if $v_Y=w_Y$ then $\eta\equiv 1$ and $f^\eta=f_{\rhote}:=E_{\rhote}[Y\mid X]$.
\end{proposition}

\begin{proof}
Under target shift, for any integrable $h$,
\begin{align*}
    E_{\rhotr}\!\big[v_Y(Y)h(X,Y)\big]
&=E_{\rhotr}\!\big[w_Y(Y)\eta(Y)h(X,Y)\big] \\
&=E_{\rhote}\!\big[\eta(Y)h(X,Y)\big],
\end{align*}
which gives \eqref{eq:Rv_as_test} by taking $h(X,Y)=(f(X)-Y)^2$.
For the minimizer, condition on $X=x$:
\[
E_{\rhote}\!\big[\eta(Y)(f(x)-Y)^2\mid X=x\big]
=\psi(x)\,E_{\rho^\eta}\!\big[(f(x)-Y)^2\mid X=x\big],
\]
and since $\psi(x)>0$, the pointwise minimizer is the conditional mean under $\rho^\eta$, yielding
\eqref{eq:induced_regression_short}.
\end{proof}
\end{comment}

To state finite-sample guarantees we assume a Bernstein-type conditional moment bound for the random weights
appearing in operator concentration (compare Assumption~\ref{ass:target_shift_weights}).

\begin{assumption}[Incorrect-weight moment condition]\label{ass:incorrect_target_shift}
Let $w_Y=d\rhote_Y/d\rhotr_Y$ and let $v_Y:Y\to(0,\infty)$. There exist constants $V_Y>0$ and $\gamma_Y>0$ such that
for all integers $m\ge 2$,
\begin{equation}\label{ass:moment_upper_bound}
\sup_{x\in X}\ \int \frac{v_Y^{m}(y)}{w_Y(y)}\,d\rhote_Y(y\mid x)
\;\le\; \frac{1}{2}\,m!\,V_Y^{m-2}\gamma_Y^2.
\end{equation}
Moreover, for $m = 1$,
\begin{equation}\label{ass:psi_lower_bound}
\inf_{x\in X}\psi(x) = \inf_{x\in X}\ \int \frac{v_Y(y)}{w_Y(y)}\,d\rhote(y\mid x) > 0.
\end{equation}
\end{assumption}

Now we are ready to state the main theorem of this section. See Appendix~\ref{proof:target_shift_incorrect} for the proof.

\begin{theorem}[W-KRR under Target Shift with Incorrect Weights]\label{main_imperfect_ts}
Let $\rhote$ and $\rho'$ be distributions on $X \times [-M,M]$ satisfying target shift and Assumptions \ref{source_condition}, \ref{ass:eff_dim}, \ref{ass:incorrect_target_shift}.
Let $n$ and $\lambda$ satisfy $\lambda \leq \|T\|$ and
\begin{equation}\label{opt_reg_imperfect_ts}
    \lambda = \left(\frac{8DE_{s}(\sqrt{V_Y}+\gamma_Y)\log\left(\frac{6}{\delta} \right)}{\sqrt{n}}\right)^{\frac{2}{2r+s}}
\end{equation}
for $\delta \in (0,1),$  and $D = \max\{1,1/\inf \psi(x)\}$.
Then, with probability greater than $1-\delta,$
\begin{equation}\label{generalization_bound_imperfect_ts}
    \begin{aligned}
    \|f_{\mathbf{z},\lambda}^{\eta}-f_{\mathcal{H}}\|_{\rhote_X}\leq & C \left(\frac{8DE_s(\sqrt{V_Y}+\gamma_Y)\log\left(\frac{6}{\delta} \right)}{\sqrt{n}}\right)^{\frac{2r}{2r+s}} \\
    &+4\left\|f_{\mathcal{H}}^{\eta}-f_{\mathcal{H}}\right\|_{\rhote_X},
    \end{aligned} 
\end{equation}
where $C = 6 D\left(M+R\right).$
\end{theorem}

Bound~\eqref{generalization_bound_imperfect_ts} identifies two structurally distinct
sources of error. The first term decays at
the minimax rate $\mathcal{O}(n^{-r/(2r+s)})$ established in
Theorem~\ref{IW_KRR_TarS}. The second term,
$\|\phi/\psi - f_{\rho^{te}}\|_{\rho^{te}_X}$, is a
population-level bias determined entirely by the discrepancy between
$v_Y$ and $w_Y$: it is independent of $n$, independent of $\mathcal{H}$, and
vanishes if and only if $v_Y = w_Y,$ $\rhotr_Y$-almost surely.

This phenomenon stands in sharp contrast to the covariate shift setting.
Under covariate shift with misspecified input-space weights, \citet{gogolashvili2023importance} establish that the induced
bias takes the form $\|f'_{\mathcal{H}} - f_{\mathcal{H}}\|_{\rho^{te}_X}$, where $f'_{\mathcal{H}}$ and
$f_{\mathcal{H}}$ are projections of the \emph{same} function $f_{\rho^{te}}$ onto
$\mathcal{H}$ under two different inner products. As the capacity of $\mathcal{H}$
increases, both projections converge to $f_{\rho^{te}}$ and the bias
vanishes, providing a way for bypassing importance
weighting in high-capacity models. Under target shift, the bias
$\|f^{\eta}_{\mathcal{H}}-f_{\mathcal{H}}\|_{\rho^{te}_X}$ represents a
discrepancy between two distinct population regression functions (projected on RKHS), not
between two projections of the same one.  Enlarging $H$ refines the
accuracy with which the estimator approximates $f^\eta$, but leaves the
fundamental misalignment between $f^\eta$ and $f_{\rho^{te}}$ intact.
Accurate estimation of $w_Y$ is therefore necessary under target shift
regardless of model capacity.

\
\begin{figure}[t]
\centering
\input{Figures/target_shift}
\caption{Irreducible bias under target shift with misspecified weights.
Incorrect weights $v_Y\neq w_Y$ induce a tilted conditional distribution
$\rho^{\eta}(dy\mid x)\propto \eta(y)\rhote(dy\mid x)$ and hence an induced regression function
$f^{\eta}=\phi/\psi$ different from the desired $f_{\rhote}$. The estimator concentrates around $f^{\eta}_{\mathcal{H}}$, so the
gap $\|f^{\eta}_{\mathcal{H}}-f_{\mathcal{H}}\|_{\rhote_X}$ persists even as $n\to\infty$.}
\label{fig:target_shift_illustration}
\end{figure}

\section{Classification under Target Shift}\label{sec:binary_classification}

So far we studied regression with $Y\subset\mathbb{R}$. We now turn to \emph{binary classification} with
$Y=\{-1,+1\}$ and show how the regression guarantees obtained under target shift immediately yield fast
rates for classification via plug-in (sign) rules.

\paragraph{From regression to classification.}
Given a real-valued score function $f:X\to\mathbb{R}$, we predict $\mathrm{sgn}(f(x))$. The corresponding
misclassification risk is
\[
\mathcal R(f):=\rhote\big(\mathrm{sgn}(f(X))\neq Y\big).
\]
Let
\[
f_{\rhote}(x)=\rhote(Y=1\mid x)-\rhote(Y=-1\mid x)
\]
denote the regression function, whose sign is the Bayes classifier and satisfies
$\inf_f \mathcal R(f)=\mathcal R(f_\rho)$.

A standard calibration inequality bounds the excess classification risk by the regression error:
\begin{equation}\label{eq:calibration_basic}
\mathcal R(f)-\mathcal R(f_{\rhote})\ \le\ \|f-f_{\rhote}\|_{\rhote_X}.
\end{equation}
Moreover, if the distribution has low noise near the decision boundary, one obtains a sharper conversion.

\paragraph{Tsybakov noise (margin) condition.}
Assume that for some $l\ge 0$ and $B_l>0$,
\begin{equation}\label{eq:tsybakov_margin}
\rhote_X\!\left(\left\{x\in X:\ |f_{\rhote}(x)|\le \Delta\right\}\right)\ \le\ B_l\,\Delta^l,
\qquad \forall \Delta\in[0,1].
\end{equation}
Under \eqref{eq:tsybakov_margin}, there exists a constant $C_l>0$ (depending only on $B_l$) such that
for all measurable $f$,
\begin{equation}\label{eq:calibration_margin}
\mathcal R(f)-\mathcal R(f_{\rhote})\ \le\ C_l\,\|f-f_{\rhote}\|_{\rhote_X}^{\,\frac{2(l+1)}{l+2}}.
\end{equation}
Inequalities of the form \eqref{eq:calibration_basic}--\eqref{eq:calibration_margin} are classical; see,
e.g., \citet{bartlett2006convexity,bauer2007regularization,yao2007early}.

\paragraph{Classification rates for IW-KRR under target shift.}
We now combine \eqref{eq:calibration_margin} with the regression bound from Theorem~\ref{IW_KRR_TarS}.
In the realizable case $f_\rho\in\mathcal H$ (hence $f_{\mathcal H}=f_\rho$) and with $M=1$, the target-shift
analysis yields the following classification guarantee.

\begin{theorem}[Binary classification under target shift]\label{thm:classification_ts}
Assume the conditions of Theorem~\ref{IW_KRR_TarS} with $Y=\{-1,+1\}$ (so $M=1$), and suppose that
$f_{\rhote}\in\mathcal H$. In addition, assume the Tsybakov noise condition \eqref{eq:tsybakov_margin}.
Let $\delta\in(0,1)$ and choose $\lambda$ as in \eqref{opt_reg_ts}. Then for $n$ large enough so that
$\lambda\le \|T\|$, with probability at least $1-\delta$,
\begin{equation}\label{eq:classification_bound_ts_clean}
\mathcal R\!\big(f^{\rm IW}_{\mathbf z,\lambda}\big)-\mathcal R(f_{\rhote})
\ \le\
C\left(\frac{8E_s(\sqrt{W_Y}+\sigma_Y)\log\!\left(\frac{6}{\delta}\right)}{\sqrt{n}}\right)^{\frac{4r(l+1)}{(2r+s)(l+2)}},
\end{equation}
where $C = C_l (3+3R)^{\frac{2(l+1)}{l+2}}.$ 
\end{theorem}

\begin{proof}
By Theorem~\ref{IW_KRR_TarS} (with $M=1$ and $f_{\mathcal H}=f_{\rhote}$), we control
$\|f^{\rm IW}_{\mathbf z,\lambda}-f_{\rhote}\|_{\rhote_X}$ with high probability. Plugging this bound into the
calibration inequality \eqref{eq:calibration_margin} yields \eqref{eq:classification_bound_ts_clean}.
\end{proof}

Theorem~\ref{thm:classification_ts} shows that the regression analysis developed for target shift directly yields
fast classification rates under the margin condition \eqref{eq:tsybakov_margin}. The exponent smoothly interpolates
between the worst-case regime $l=0$ (where \eqref{eq:calibration_basic} gives a linear conversion) and low-noise
regimes $l\to\infty$ (where the conversion becomes nearly quadratic). The dependence on the shift severity enters
only through the same weight-moment quantities $(W_Y,\sigma_Y)$ as in the regression bound.

\subsection{Bias under target shift in binary classification}\label{sec:incorrect_weights_binary}

The bias mechanism from Section~\ref{sec:incorrect_weights} becomes particularly transparent when
$Y \in \{-1,+1\}$. Let $w_Y = d\rhote_Y/d\rhotr_Y$ be the true target-shift weight, let $v_Y$ be an
arbitrary (possibly incorrect) weight function, and define the relative weight errors
$\eta_+ := v_Y(+1)/w_Y(+1)$ and $\eta_- := v_Y(-1)/w_Y(-1)$.
Writing $p(x) := \rhote(Y=+1 \mid x)$, the induced regression function from
\eqref{eq:induced_regression_short} reduces to
\begin{equation}\label{eq:induced_regression_binary}
f^\eta(x) = \frac{\eta_+ p(x) - \eta_-(1-p(x))}{\eta_+ p(x) + \eta_-(1-p(x))}.
\end{equation}
The entire misspecification is captured by the two scalars $\eta_+$ and $\eta_-$.
In particular, the bias $\|f^\eta - f_{\rhote}\|_{\rhote_X}$ vanishes if and only if
$\eta_+ = \eta_-$, i.e., whenever $v_Y$ preserves the ratio $w_Y(+1)/w_Y(-1)$,
even if the individual weights are incorrect. When $\eta_+ \neq \eta_-$, the
induced decision boundary $f^\eta(x) = 0$ corresponds to $p(x) = \eta_-/(\eta_+ + \eta_-)$
rather than the Bayes-optimal threshold $p(x) = 1/2$, so the ratio $\eta_+/\eta_-$
acts as an implicit cost asymmetry.

A natural special case is $v_Y \equiv 1$ (no reweighting). Writing
$\pi^{tr} := \rhotr_Y(+1)$ and $\pi^{te} := \rhote_Y(+1)$, we have
$w_Y(+1) = \pi^{te}/\pi^{tr}$ and $w_Y(-1) = (1-\pi^{te})/(1-\pi^{tr})$,
so $\eta_+ = \pi^{tr}/\pi^{te}$ and $\eta_- = (1-\pi^{tr})/(1-\pi^{te})$.
In this case the induced score simplifies to
\begin{equation}\label{eq:unweighted_is_training_score}
f^\eta(x) = 2\,\rhotr(Y=+1 \mid x) - 1,
\end{equation}
i.e., an unweighted method learns the \emph{training} posterior score rather
than the test-optimal one. If an unweighted estimator yields
$\hat{s}(x) \approx f^\eta(x)$, the Bayes score $f_{\rhote}(x)$ can be
recovered via the fractional-linear transform
\begin{equation}\label{eq:posthoc_correction}
f_{\rhote}(x) = \frac{(\eta_- - \eta_+) + \hat{s}(x)(\eta_+ + \eta_-)}{(\eta_+ + \eta_-) + \hat{s}(x)(\eta_- - \eta_+)},
\end{equation}
which depends only on the class proportions $(\pi^{tr}, \pi^{te})$.
This is precisely the classical Bayes-rule recalibration of the posterior
\citep{saerens2002adjusting, lipton2018detecting}, expressed in the score
parametrization $\hat{s}(x) = 2\,\rhotr(Y=+1 \mid x) - 1$. In practice,
$\pi^{tr}$ is known from the training labels and $\pi^{te}$ can be estimated
from unlabeled test data using label-shift estimators
\citep{saerens2002adjusting, lipton2018detecting}.

\section{Simulations}\label{sec:simulations}
We present a simple simulation illustrating the key qualitative 
predictions of our theory. A comprehensive empirical study is 
beyond the scope of this work; our aim here is to confirm 
the contrasting roles of importance weighting under covariate 
and target shift.

We validate our theoretical findings through simulations comparing covariate and target shift scenarios, focusing on three key predictions: for high-capacity models under covariate shift, IW correction is unnecessary; for low-capacity models under covariate shift, IW correction is beneficial; and under target shift, IW correction is beneficial regardless of model capacity.

We consider one-dimensional regression with $f_{\rhote}(x) = x^3$ and output noise $y = f_{\rhote}(x) + \varepsilon$ where $\varepsilon \sim \mathcal{N}(0, 0.3^2)$. Training and test sets contain $n=200$ points each.

Under covariate shift, we set $x \sim \mathcal{N}(0.8, 0.5^2)$ at training and $x \sim \mathcal{N}(0, 0.35^2)$ at test. Under target shift, we set $y \sim \mathcal{N}(0, 0.5^2)$ at training and $y \sim \mathcal{N}(1.5, 0.3^2)$ at test, with $x = (y+\varepsilon)^{1/3}$ where $\varepsilon \sim \mathcal{N}(0, 0.3^2)$.

\begin{figure}[t]
  \vspace{-10pt}
  \subfloat[Regression under covariate shift]{\includegraphics[width=0.5\textwidth]{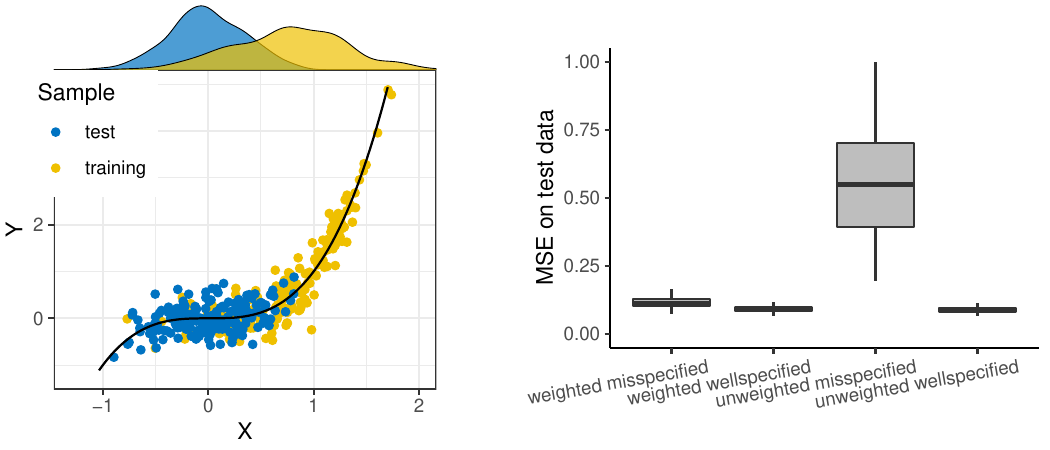}}
  \subfloat[Regression under target shift]{\includegraphics[width=0.5\textwidth]{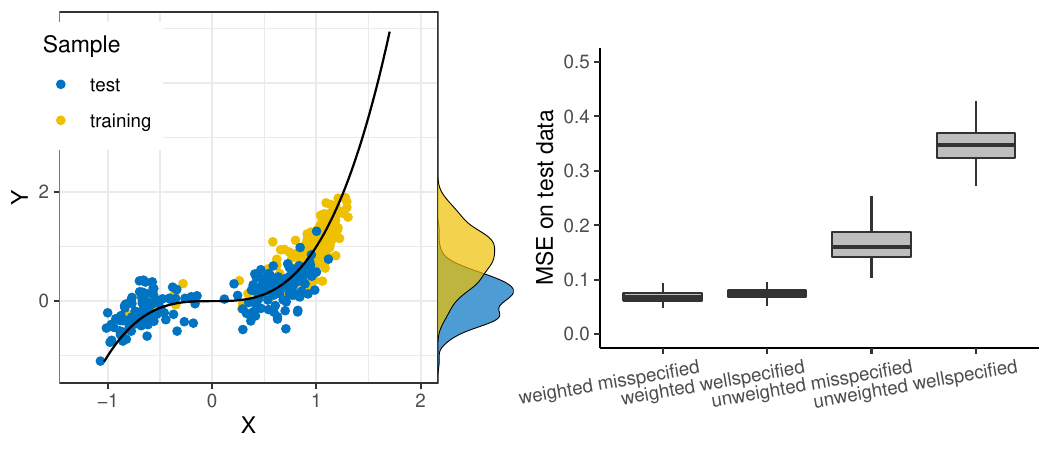}}
  \caption{Performance comparison for different shift scenarios. Left panels show data and regression function. Right panels show MSE boxplots over 200 replications. (a)~Covariate shift: well-specified unweighted model performs comparably to IW. (b)~Target shift: IW correction is essential regardless of capacity.}
\label{CovS_vs_TarS}
\end{figure}

We use polynomial kernels with degree~2 (misspecified) and degree~3 (well-specified), and Figure~\ref{CovS_vs_TarS} shows mean squared error (MSE) over 200 replications.

Figure~\ref{CovS_vs_TarS}a confirms that under covariate shift, well-specified unweighted models match IW-corrected performance, consistent with \citet{gogolashvili2023importance}. In contrast, Figure~\ref{CovS_vs_TarS}b shows that under target shift, omitting IW correction yields significantly higher MSE regardless of model specification, validating our theoretical prediction that IW correction is essential for target shift.

\section{Conclusion}

We studied importance-weighted kernel ridge regression under target shift, 
establishing minimax-optimal convergence rates of order $\mathcal{O}((W/n)^{r/(2r+s)})$ 
under relaxed Bernstein-type moment conditions on the output-space weights. 
A key finding is that the distributional mismatch enters the bound only 
through scalar constants $(W_Y, \sigma_Y)$, leaving the smoothness-driven 
exponent $n^{-r/(2r+s)}$ unchanged from the no-shift setting. This 
contrasts with covariate shift, where the weights interact with the 
input-space geometry and can degrade the effective dimension. Notably, 
when the weights are uniformly bounded, both shift types yield the same 
minimax rate, but the mechanisms differ: under covariate shift, heavy-tailed 
weights cause more severe deterioration, whereas under target shift the 
weight bound acts as a scalar prefactor.

We also analyzed the effect of misspecified importance weights. Under target 
shift, incorrect weights induce an irreducible bias 
$\|f^{\eta}_{\mathcal{H}} - f_{\mathcal{H}}\|_{\rhote_X}$ representing a discrepancy between 
two distinct projected regression functions. Unlike covariate shift, where increasing 
model capacity eliminates the bias from weight misspecification, this bias 
persists regardless of the expressiveness of $\mathcal{H}$. Accurate 
estimation of the label-marginal ratio $w_Y$ is therefore essential under 
target shift.

Several directions remain open. A natural next step is to combine the 
misspecified-weight bounds developed here with finite-sample guarantees for 
label-shift estimators, yielding end-to-end rates for IW-KRR with estimated 
weights. It would also be of interest to investigate how this covariate-versus-target 
shift dichotomy manifests for scalable kernel approximations such as Nyström 
and random feature methods, where the interplay between approximation capacity 
and weight quality may differ from the exact kernel setting.

%Bibliography
\bibliographystyle{plainnat}
\bibliography{references}

\appendix

\section*{Appendix}
In this appendix, we provide proofs of our main results: 
Theorem~\ref{IW_KRR_TarS} (optimal rates under target shift), 
Theorem~\ref{theo:lower_bound_ts} (minimax lower bound), and 
Theorem~\ref{main_imperfect_ts} (incorrect weights). We begin by 
collecting standard auxiliary results from the kernel learning 
literature.

\section{Auxiliary Results}

The following Bernstein inequality for Hilbert space-valued random variables is from \citet[Proposition~2]{caponnetto2007optimal}:

\begin{proposition}[Bernstein Inequality]\label{bernstein}
Let $(Z, \rho)$ be a probability space and let $\xi$ be a random variable on $Z$ taking values in a real separable Hilbert space $H.$ Assume there exist positive constants $L$ and $\sigma$ such that
$$
\mathbb{E}\left[\|\xi-\mathbb{E}[\xi]\|_{H}^{m}\right] \leq \frac{1}{2} m ! \sigma^{2} L^{m-2}, \quad \forall m \geq 2.
$$
Then, for any $\delta \in (0,1]$,
$$
\left\|\frac{1}{n} \sum_{i=1}^{n} \xi\left(z_{i}\right)-\mathbb{E}[\xi]\right\|_{H} \leq \frac{2L\log(2/\delta)}{n}+\sqrt{\frac{2\sigma^{2}\log(2/\delta)}{n}}
$$
with probability at least $1-\delta$.
\end{proposition}

The following bound on approximation error is standard (see \citet{smale2007learning}):

\begin{proposition}[Approximation Error]\label{approximation_error}
Let $f_{\mathcal{H}}$ satisfy Assumption~\ref{source_condition} for some $r>0,$ and define
$f_{\lambda} = (T+\lambda)^{-1}Lf_{\mathcal{H}}.$
Then
\[
\left\|f_{\lambda}-f_{\mathcal{H}}\right\|_{\rhote_X} \leq \lambda^{r}\left\|L^{-r} f_{\mathcal{H}}\right\|_{\rhote_X} \leq \lambda^r R \quad \text{for } r \leq 1.
\]
Furthermore, for $r>0.5$,
$\left\|f_{\lambda}\right\|_{\mathcal{H}} \leq \left\|L^{-r} f_{\mathcal{H}}\right\|_{\rhote_X} \leq R.$
\end{proposition}

The following result on mixed operators is from \citet[Proposition~8]{rudi2017generalization}:

\begin{proposition}[Mixed Operators]\label{mixed_operators}
For any $\lambda>0$ and bounded self-adjoint positive operators $A, B$ on separable Hilbert space $\mathcal{H},$
$$
\left\|(A+\lambda I)^{-1 / 2} B^{1 / 2}\right\| \leq \left\|(A+\lambda I)^{-1 / 2} (B+\lambda)^{1 / 2}\right\| \leq(1-\mu)^{-1 / 2}
$$
with $\mu=\mu_{\max }\left[(B+\lambda I)^{-1 / 2}(B-A)(B+\lambda I)^{-1 / 2}\right]$.
\end{proposition}

\subsection{Minimax lower bound under target shift}\label{proof:lower_bound_my_proof}

\paragraph{Construction of the distributions.}
Let $(X,\nu)$ be the input probability space and fix the \emph{test} marginal $\rhote_X \equiv \nu$.
Consider three pairwise disjoint measurable subsets $A,B,C \subseteq [-M,M]$ and let $q_A,q_B,q_C$ be the
corresponding uniform densities. Fix $c:=1/4$ and define the test and training $Y$-marginals by
\begin{align}\label{marginals_continuous}
\rhote_Y(y) &= c\,q_A(y)+c\,q_B(y)+(1-2c)\,q_C(y), \nonumber\\
\rhotr_Y(y) &= \frac{c}{W}\,q_A(y)+\frac{c}{W}\,q_B(y)+\Bigl(1-\frac{2c}{W}\Bigr)\,q_C(y).
\end{align}
The pointwise importance weight satisfies
\begin{equation}\label{weights_continuous}
w(y)=\frac{d\rhote_Y}{d\rhotr_Y}(y)=
\begin{cases}
W, & y\in A\cup B,\\[2mm]
\dfrac{1-2c}{1-2c/W}\le 1, & y\in C,
\end{cases}
\end{equation}
and hence $w(y)\le W$ everywhere. In particular, the training distribution assigns only
\begin{equation}\label{informative_mass}
\rhotr_Y(A\cup B)=\frac{2c}{W}=\frac{1}{2W}
\end{equation}
mass to the \emph{informative} label region $A\cup B$.

\paragraph{Construction of the hypothesis space.}
Let
\[
\ell_n:=\Bigl(\frac{W}{n}\Bigr)^{\frac{2r}{2r+s}}
\qquad\text{and}\qquad
m:=\Bigl\lfloor C_0\Bigl(\frac{W}{n}\Bigr)^{-\frac{s}{2r+s}}\Bigr\rfloor,
\]
where $C_0>0$ will be chosen later and we assume $n$ is large enough so that $m\ge 8$.
Consider the class
\[
\mathcal{E}:=\Bigl\{f_a(x)=\sqrt{\frac{\ell_n}{m}}\sum_{i=1}^m \alpha_i \mu_i^{-r}e_i(x)\,:\,
a=(\alpha_1,\dots,\alpha_m)\in\{0,1\}^m\Bigr\}.
\]
Recall that $(e_i)_{i\ge 1}$ is an orthonormal system in $L^2(\rhote_X)=L^2(\nu)$; moreover, we take $e_i$ to be
mean-zero, i.e.\ $\int e_i\,d\nu=0$ for all $i\ge 1$ (this is possible whenever $\nu$ is not a Dirac measure).
Then, using $\mu_i\ge A i^{-1/s}$,
\[
\|f_a\|_{\rhote_X}^2
=\frac{\ell_n}{m}\sum_{i=1}^m \mu_i^{-2r}
\le \frac{\ell_n}{m}\sum_{i=1}^m A^{-2r} i^{2r/s}
\le \ell_n A^{-2r} m^{2r/s}.
\]
With $C_0:=R^{s/r}A^s$ this yields $\|f_a\|_{\rhote_X}\le R$.

By the Varshamov--Gilbert lemma \citep[Lemma 2.9]{tsybakov2009introduction}, there exists a subset
$\{a_0,\dots,a_M\}\subset\{0,1\}^m$ with $a_0=(0,\dots,0)$ such that for all $j\neq k$,
\begin{equation}\label{VG_hamming_lower_bound}
\mathrm{Hamming}(a_j,a_k)\ge \frac{m}{8},
\end{equation}
and
\begin{equation}\label{VG_number_of_set}
M\ge 2^{m/8}.
\end{equation}
Define the regression functions
\[
f_j(x):=(L^r f_{a_j})(x),\qquad j=0,\dots,M.
\]
Since $L^r(\mu_i^{-r}e_i)=e_i$, we have the explicit form
\[
f_j(x)=\sqrt{\frac{\ell_n}{m}}\sum_{i=1}^m \alpha_i^{(j)} e_i(x),
\qquad\text{and hence}\qquad
\|f_j\|_{\rhote_X}^2\le \ell_n.
\]

\paragraph{Separation.}
For $j\neq k$,
\[
\|f_j-f_k\|_{\rhote_X}^2
=\frac{\ell_n}{m}\sum_{i=1}^m(\alpha_i^{(j)}-\alpha_i^{(k)})^2
\ge \frac{\ell_n}{m}\,\mathrm{Hamming}(a_j,a_k)
\ge \frac{\ell_n}{8},
\]
where we used \eqref{VG_hamming_lower_bound}.

\paragraph{Target-shift model.}
Let $\Psi(y):=\mathbf{1}_A(y)-\mathbf{1}_B(y)\in\{-1,0,+1\}$.
For each $j$ define the conditional distribution of $X$ given $Y=y$ via the density perturbation
\begin{equation}\label{conditional-perturbation}
\frac{d\rho_j(\cdot\mid y)}{d\nu}(x)=1+t_y^{(j)}(x),
\qquad
t_y^{(j)}(x):=\delta\,\Psi(y)\,f_j(x),
\end{equation}
where $\delta>0$ will be fixed below. Since $\int f_j\,d\nu=0$ and $\Psi(y)$ is constant in $x$,
\[
\int \frac{d\rho_j(\cdot\mid y)}{d\nu}(x)\,d\nu(x)
=1+\delta\Psi(y)\int f_j\,d\nu
=1,
\]
so $\rho_j(\cdot\mid y)$ is a probability measure. We also assume $|t_y^{(j)}(x)|\le 1/2$ $\nu$-a.s.
To verify $|t_y^{(j)}(x)|\le 1/2$, note that $f_j\in\mathcal{H}$ 
with
\[
\|f_j\|_{\mathcal{H}}^2
= \frac{\ell_n}{m}\sum_{i=1}^m \mu_i^{-1}
\lesssim \ell_n\, m^{1/s}
= \Bigl(\frac{W}{n}\Bigr)^{\frac{2r-1}{2r+s}},
\]
where we used $\mu_i^{-1}\leq A^{-1}i^{1/s}$. Since $r>1/2$, the 
exponent $(2r-1)/(2r+s)$ is strictly positive, so 
$\|f_j\|_{\mathcal{H}}\to 0$ as $n\to\infty$. The reproducing 
property gives $\|f_j\|_\infty \le \|f_j\|_{\mathcal{H}}$, 
and hence $|t_y^{(j)}(x)| = \delta|\Psi(y)||f_j(x)| 
\le \delta\|f_j\|_{\mathcal{H}} \le 1/2$ for $n$ 
sufficiently large. For $r = 1/2$, we can choose $\delta$ small enough to ensure $\|f_j\|_{\mathcal{H}} \le 1/2.$

Define the training joint laws
\[
\rhotr_j(dx,dy):=\rho_j(dx\mid y)\,\rhotr_Y(dy),
\qquad
\rhotr_0(dx,dy):=\nu(dx)\,\rhotr_Y(dy),
\]
where $f_0\equiv 0$ implies $\rho_0(\cdot\mid y)\equiv \nu$.

\paragraph{One-sample KL computation.}
Since $\rhotr_j$ and $\rhotr_0$ share the same $Y$-marginal $\rhotr_Y$, their Radon--Nikodym derivative factorizes as
\[
\frac{d\rhotr_j}{d\rhotr_0}(x,y)=\frac{d\rho_j(\cdot\mid y)}{d\nu}(x)=1+t_y^{(j)}(x).
\]
Therefore,
\begin{align}\label{kl-basic-formula}
\KL(\rhotr_j\Vert\rhotr_0)
&=\iint \log\!\Bigl(1+t_y^{(j)}(x)\Bigr)\,d\rhotr_j(x,y)\nonumber\\
&=\int \rhotr_Y(dy)\int \rho_j(dx\mid y)\,\log\!\Bigl(1+t_y^{(j)}(x)\Bigr)\nonumber\\
&=\int \rhotr_Y(dy)\int \Bigl(1+t_y^{(j)}(x)\Bigr)\log\!\Bigl(1+t_y^{(j)}(x)\Bigr)\,d\nu(x).
\end{align}

\paragraph{The $1/W$ factor.}
If $y\in C$, then $\psi(y)=0$ and hence $t_y^{(j)}\equiv 0$, so the integrand in \eqref{kl-basic-formula} vanishes.
Thus only $y\in A\cup B$ contribute and
\begin{equation}\label{kl-informative-region}
\KL(\rhotr_j\Vert\rhotr_0)
=\int_{A\cup B}\!\rhotr_Y(dy)\int \Bigl(1+t_y^{(j)}\Bigr)\log\!\Bigl(1+t_y^{(j)}\Bigr)\,d\nu.
\end{equation}

\paragraph{Bounding the integrand.}
For $|t|\le 1/2$ we have the pointwise inequality
\[
(1+t)\log(1+t)\le \frac{t^2}{1-|t|}\le 2t^2.
\]
Using $t=t_y^{(j)}(x)=\delta\psi(y)f_j(x)$ and $\psi(y)^2=1$ on $A\cup B$,
\[
\int \Bigl(1+t_y^{(j)}\Bigr)\log\!\Bigl(1+t_y^{(j)}\Bigr)\,d\nu
\le 2\delta^2\|f_j\|_{\nu}^2
\le 2\delta^2\ell_n,
\qquad y\in A\cup B.
\]
Combining with \eqref{kl-informative-region} and \eqref{informative_mass} yields the one-sample bound
\begin{equation}\label{kl-one-sample-bound}
\KL(\rhotr_j\Vert\rhotr_0)
\le \rhotr_Y(A\cup B)\cdot 2\delta^2\ell_n
=\frac{2c}{W}\cdot 2\delta^2\ell_n
=\frac{4c\delta^2}{W}\,\ell_n.
\end{equation}
Consequently, for $n$ i.i.d.\ samples,
\[
\KL(\rhotr_{n,j}\Vert\rhotr_{n,0})
=n\,\KL(\rhotr_j\Vert\rhotr_0)
\le \frac{4c\delta^2}{W}\,n\ell_n.
\]

\paragraph{Choice of parameters and application of Tsybakov's theorem.}
By \eqref{VG_number_of_set} we have $\log M \ge (m/8)\log 2$. Since
\[
\frac{n\ell_n}{W}=\Bigl(\frac{n}{W}\Bigr)^{\frac{s}{2r+s}},
\qquad
m=\Bigl\lfloor C_0\Bigl(\frac{n}{W}\Bigr)^{\frac{s}{2r+s}}\Bigr\rfloor,
\]
it follows that for $n$ large enough,
\[
\KL(\rhotr_{n,j}\Vert\rhotr_{n,0})
\le \Bigl(\frac{32c\delta^2}{C_0\log 2}\Bigr)\,\log M
=: \alpha \log M.
\]
Choosing $\delta$ such that $\alpha\le 1/8$, Theorem~2.5 in
\citet{tsybakov2009introduction} implies that for any estimator $\hat f_n$ measurable w.r.t.\ the training sample,
\[
\sup_{j\in\{0,\dots,M\}}
\mathbb{P}_{\rhotr_{n,j}}\!\left(\|\hat f_n-f_j\|_{\rhote_X}^2 \ge \frac{\ell_n}{16}\right)
\ge c_1,
\]
for a universal constant $c_1>0$. By Markov's inequality, for any estimator $\hat f_n$,
\[
\mathbb{E}_{\rhotr_{n,j}}\|\hat f_n-f_j\|_{\rhote_X}^2
\ge \frac{\ell_n}{16} \cdot \mathbb{P}_{\rhotr_{n,j}}\!\left(\|\hat f_n-f_j\|_{\rhote_X}^2 \ge \frac{\ell_n}{16}\right).
\]
Taking the infimum over $\hat f_n$ and supremum over $j$, we obtain
\[
\inf_{\hat f_n}\ \sup_{j\in\{0,\dots,M\}}
\mathbb{E}_{\rhotr_{n,j}}\|\hat f_n-f_j\|_{\rhote_X}^2
\gtrsim \ell_n
=\Bigl(\frac{W}{n}\Bigr)^{\frac{2r}{2r+s}}.
\]
By Jensen's inequality, we obtain the lower bound for the non-squared norm.

\subsection{Proof of Theorem~\ref{main_imperfect_ts}: W-KRR under Target Shift with Incorrect Weights}\label{proof:target_shift_incorrect}

Throughout this proof we write $T_{\mathbf{z}} := S^{\top}_{\mathbf{x}}M_{\mathbf{v}}S_{\mathbf{x}}$,
$g_{\mathbf{z}} := S^{\top}_{\mathbf{x}}M_{\mathbf{v}}\mathbf{y}$, where $M_{\psi}$ for the multiplication
operator $M_{\psi}f := f\psi$, and $\mathbf{v} = (v_Y(y_1),\dots,v_Y(y_n))^{\top}$.
We also let $f_{\lambda} := (T+\lambda)^{-1}Tf_{\mathcal{H}}$ denote the population regularised solution
associated with the correct test distribution.

\paragraph{Step 1: Excess risk decomposition.}

Adding and subtracting $f_{\lambda}$ and exploiting the operator identity
$(T_{\mathbf{z}}+\lambda)^{-1}(T_{\mathbf{z}}+\lambda) = I$, the excess risk
$f_{\mathbf{z},\lambda}^v - f_{\mathcal{H}}$ decomposes into three terms:
\begin{equation}\label{TS_excess_risk}
    \begin{aligned}
    f_{\mathbf{z},\lambda}^v-f_{\mathcal{H}}
    \;&=\;
    \underbrace{%
        \left(T_{\mathbf{z}}+\lambda\right)^{-1}
        \Bigl[\left(g_{\mathbf{z}}-L\phi\right)
              +\left(LM_{\psi}-T_{\mathbf{z}}\right)f_{\lambda}\Bigr]
    }_{\displaystyle\mathrm{I}:\;\text{stochastic error}}\\
    \;&+\;
    \underbrace{%
        \left(T_{\mathbf{z}}+\lambda\right)^{-1}
        \Bigl[L\left(\phi-f_{\rhote}\right)
              +\left(T-LM_{\psi}\right)f_{\lambda}\Bigr]
    }_{\displaystyle\mathrm{II}:\;\text{weight-mismatch bias}}\\
    \;&+\;
    \underbrace{%
        f_{\lambda}-f_{\mathcal{H}}
    }_{\displaystyle\mathrm{III}:\;\text{approximation error}}.
    \end{aligned}
\end{equation}
We bound each term in turn.

\paragraph{Step 2: Stochastic error (Term~I).}
We have
\[
\mathrm{I}:=(T_{\mathbf z}+\lambda I)^{-1}u,
\quad
u:=(g_{\mathbf z}-L\phi)+(LM_\psi-T_{\mathbf z})f_\lambda.
\]
Since $\|h\|_{\rhote_X}=\|T^{1/2}h\|_{\mathcal H}$ for $h\in\mathcal H$, we have
\[
\|\mathrm{I}\|_{\rhote_X}
=\big\|T^{1/2}(T_{\mathbf z}+\lambda I)^{-1}u\big\|_{\mathcal H}.
\]
Insert $(T+\lambda I)^{\pm 1/2}$ and $(LM_\psi+\lambda I)^{\pm 1/2}$:
\[
T^{1/2}(T_{\mathbf z}+\lambda I)^{-1}
=
T^{1/2}(T+\lambda I)^{-1/2}\,
(T+\lambda I)^{1/2}(LM_\psi+\lambda I)^{-1/2}\,
(LM_\psi+\lambda I)^{1/2}(T_{\mathbf z}+\lambda I)^{-1}.
\]
The first factor satisfies $\|T^{1/2}(T+\lambda I)^{-1/2}\|\le 1$, hence
\begin{equation}\label{eq:split_termI}
\|\mathrm{I}\|_{\rhote_X}
\le
\|(T+\lambda I)^{1/2}(LM_\psi+\lambda I)^{-1/2}\|\,
\|(LM_\psi+\lambda I)^{1/2}(T_{\mathbf z}+\lambda I)^{-1}u\|_{\mathcal H}.
\end{equation}

\smallskip\noindent
\textbf{Step 2a: the $(1-\mu)^{-1/2}$ factor.}
Define
\[
\mu:=\mu_{\max}\!\Big((T+\lambda I)^{-1/2}(T-LM_\psi)(T+\lambda I)^{-1/2}\Big),
\]
by lemma~\ref{mixed_operators} we have
\begin{equation}\label{eq:mu_factor_termI}
\|(T+\lambda I)^{1/2}(LM_\psi+\lambda I)^{-1/2}\|\le (1-\mu)^{-1/2}.
\end{equation}

\smallskip\noindent
\textbf{Step 2b: Neumann-series bound and the $(1-S_1')^{-1}$ factor.}
Let
\[
\Delta_\psi:=(LM_\psi+\lambda I)^{-1/2}(LM_\psi-T_{\mathbf z})(LM_\psi+\lambda I)^{-1/2}.
\]
Then
\[
T_{\mathbf z}+\lambda I=(LM_\psi+\lambda I)^{1/2}(I-\Delta_\psi)(LM_\psi+\lambda I)^{1/2},
\]
hence
\[
(LM_\psi+\lambda I)^{1/2}(T_{\mathbf z}+\lambda I)^{-1}=(I-\Delta_\psi)^{-1}(LM_\psi+\lambda I)^{-1/2}.
\]
Therefore,
\[
\|(LM_\psi+\lambda I)^{1/2}(T_{\mathbf z}+\lambda I)^{-1}u\|_{\mathcal H}
\le \|(I-\Delta_\psi)^{-1}\|\,\|(LM_\psi+\lambda I)^{-1/2}u\|_{\mathcal H}.
\]
Since $\|\Delta_\psi\|\le \|\Delta_\psi\|_{\mathrm{HS}}=:S_1'$, whenever $S_1'<1$ the Neumann series yields
\begin{equation}\label{eq:neumann_termI}
\|(I-\Delta_\psi)^{-1}\|\le \frac{1}{1-\|\Delta_\psi\|}\le \frac{1}{1-S_1'}.
\end{equation}

\smallskip\noindent
\textbf{Step 2c: converting $(LM_\psi+\lambda I)^{-1/2}$ to $(T+\lambda I)^{-1/2}$.}
From $LM_\psi+\lambda I\succeq (1-\mu)(T+\lambda I)$ we also obtain
\[
(LM_\psi+\lambda I)^{-1/2}\preceq (1-\mu)^{-1/2}(T+\lambda I)^{-1/2},
\]
and thus
\begin{equation}\label{eq:mu_second_termI}
\|(LM_\psi+\lambda I)^{-1/2}u\|_{\mathcal H}
\le (1-\mu)^{-1/2}\,\|(T+\lambda I)^{-1/2}u\|_{\mathcal H}.
\end{equation}
Finally, by the triangle inequality and the definition of $u$,
\[
\|(T+\lambda I)^{-1/2}u\|_{\mathcal H}
\le
\underbrace{\|(T+\lambda I)^{-1/2}(g_{\mathbf z}-L\phi)\|_{\mathcal H}}_{=:S_2'}
+
\underbrace{\|(T+\lambda I)^{-1/2}(LM_\psi-T_{\mathbf z})f_\lambda\|_{\mathcal H}}_{=:S_3'}.
\]

\smallskip\noindent
Combining \eqref{eq:split_termI}, \eqref{eq:mu_factor_termI}, \eqref{eq:neumann_termI}, and
\eqref{eq:mu_second_termI} yields
\[
\|\mathrm{I}\|_{\rhote_X}
\le (1-\mu)^{-1/2}\cdot \frac{1}{1-S_1'}\cdot (1-\mu)^{-1/2}\cdot (S_2'+S_3')
=
\frac{1}{1-\mu}\left(\frac{S_2'+S_3'}{1-S_1'}\right),
\]
where
$$
\begin{gathered}
S'_{2} :=\left\|(T+\lambda)^{-\frac{1}{2}}\left(g_{\mathbf{z}}-L\phi\right)\right\|_{\mathcal{H}}, \\
S'_{3} :=\left\|(T +\lambda)^{-\frac{1}{2}}\left(LM_{\psi}-T_{\mathbf{z}}\right)f_{\lambda}\right\|_{\mathcal{H}}, \\
S'_{1} :=\left\|(LM_{\psi}+\lambda)^{-\frac{1}{2}}\left(LM_{\psi}-T_{\mathbf{z}}\right)(LM_{\psi}+\lambda)^{-\frac{1}{2}}\right\|_{\mathrm{HS}},
\end{gathered}
$$
and
$$
\mu = \mu_{\max }\left((T+\lambda I)^{-1 / 2}(T-LM_{\psi})(T+\lambda I)^{-1 / 2}\right).
$$   
One can easily show that $\mu \leq 1-\inf \psi,$ therefore
\begin{equation}\label{ts_stoch_raw}
\|\text{I} \|_{\rhote_X} \leq D\left(\frac{S'_2+S'_3}{1-S'_1} \right).
\end{equation}
We apply Proposition~\ref{bernstein} to the centred Hilbert-space-valued random variables
\begin{align*}
\xi_{1}(z) &:= (T+\lambda)^{-\frac{1}{2}} v_Y(y)\,
               K_{x}\!\left\langle K_{x},\,\cdot\,\right\rangle_{\mathcal{H}}
               (T+\lambda)^{-\frac{1}{2}}, \\
\xi_{2}(z) &:= (T+\lambda)^{-\frac{1}{2}} v_Y(y)\,K_{x}\,y, \\
\xi_{3}(z) &:= (T+\lambda)^{-\frac{1}{2}} v_Y(y)\,K_{x}\,f_{\lambda}(x).
\end{align*}
Under Assumption~\ref{ass:incorrect_target_shift}, one verifies that the Bernstein constants are
\begin{align*}
    L'_1 = 2 \frac{V_Y}{\lambda}  ,& \quad \sigma'_1 = 2 \sqrt{\frac{\mathcal{N}(\lambda)}{\lambda}} \gamma_Y , \\
    L'_2 = 2 \frac{M V_Y}{\sqrt{\lambda}} , &\quad \sigma'_2 = 2 M \gamma_Y  \sqrt{\mathcal{N}(\lambda)}, \\
    L'_3 = 2  \frac{\|f_{\lambda}\|_{\mathcal{H}} V_Y}{\sqrt{\lambda}}, &\quad \sigma'_3 = 2 \|f_{\lambda}\|_{\mathcal{H}} \gamma_Y \sqrt{\mathcal{N}(\lambda)}.
\end{align*}
Under the regularisation condition $n\lambda \geq 16(V_Y+\gamma_Y^2)\mathcal{N}(\lambda)D^2\log^2(6/\delta)$,
Proposition~\ref{bernstein} yields $S'_1 \leq 3/4$ with probability at least $1-\delta/3$.
Substituting into \eqref{ts_stoch_raw} and using $\|f_\lambda\|_{\mathcal{H}} \leq R$
(Proposition~\ref{approximation_error}) gives, with probability at least $1-\delta$,
\begin{equation}\label{ts_I_term}
    \|\mathrm{I}\|_{\rhote_X}
    \;\leq\;
    16D(M+R)\log\!\left(\tfrac{6}{\delta}\right)
    \left(\frac{V_Y}{\sqrt{n\lambda}}
          +\gamma_Y\frac{E_s}{\sqrt{n\lambda^{s}}}\right).
\end{equation}

\begin{comment}

\paragraph{Step 3: Weight-mismatch bias (Term~II).}

We factorise Term~II by inserting $M_\psi M_\psi^{-1}$ inside the bracket:
\begin{align*}
    \mathrm{II}
    &= \left(T_{\mathbf{z}}+\lambda\right)^{-1}
       \Bigl[L\!\left(\phi-f_{\rhote}\right)
             +\left(T-LM_{\psi}\right)f_{\lambda}\Bigr] \\
    &= \left(T_{\mathbf{z}}+\lambda\right)^{-1}LM_{\psi}
       \left[\frac{\phi}{\psi}-f_{\rhote}
             +\frac{1-\psi}{\psi}\!\left(f_{\lambda}-f_{\mathcal{H}}\right)\right].
\end{align*}
It is not difficult to show that
the operator norm satisfies $\|(T_{\mathbf{z}}+\lambda)^{-1}LM_{\psi}\|\leq 4$.
Since $\inf\psi \geq 1/D$, the term $(1-\psi)/\psi$ is bounded by $D-1 \le D$, and therefore
\begin{equation}\label{ts_II_term}
    \|\mathrm{II}\|_{\rhote_X}
    \;\leq\;
    4\left\|\frac{\phi}{\psi}-f_{\rhote}\right\|_{\rhote_X}
    +5D\,\|f_{\lambda}-f_{\mathcal{H}}\|_{\rhote_X}.
\end{equation}
The first summand is the irreducible \emph{weight-mismatch bias}; it vanishes if and only if $v_Y = w_Y$.
    
\end{comment}

\paragraph{Step 3: Weight-mismatch bias (Term~II).}

Recall Term~II from \eqref{TS_excess_risk}:
\[
\mathrm{II}
=
\left(T_{\mathbf{z}}+\lambda\right)^{-1}
\Bigl[L\!\left(\phi-f_{\rhote}\right)+\left(T-LM_{\psi}\right)f_{\lambda}\Bigr].
\]
Since $f_{\mathcal{H}}$ is the $L_2(\rhote_X)$-projection of $f_{\rhote}$ onto $\overline{\mathcal{H}}$,
it satisfies the normal equation $T f_{\mathcal{H}} = L f_{\rhote}$. Hence,
\begin{align*}
L(\phi-f_{\rhote})+\left(T-LM_{\psi}\right)f_{\lambda}
&= L\phi - L f_{\rhote} + T f_\lambda - LM_\psi f_\lambda \\
&= L\phi - T f_{\mathcal{H}} + \left(T-LM_{\psi}\right)f_{\lambda} \\
&= \underbrace{\bigl(L\phi - LM_{\psi}f_{\mathcal{H}}\bigr)}_{(\star)}
   + \left(T-LM_{\psi}\right)\bigl(f_{\lambda}-f_{\mathcal{H}}\bigr).
\end{align*}
With \eqref{eq:fHv_normal_equation_in_prop}, we have $(\star)=LM_{\psi}\bigl(f_{\mathcal{H}}^{\eta}-f_{\mathcal{H}}\bigr)$, and therefore
\begin{equation}\label{eq:II_decomp}
\mathrm{II}
=
\left(T_{\mathbf{z}}+\lambda\right)^{-1}LM_{\psi}\bigl(f_{\mathcal{H}}^{\eta}-f_{\mathcal{H}}\bigr)
+
\left(T_{\mathbf{z}}+\lambda\right)^{-1}\left(T-LM_{\psi}\right)\bigl(f_{\lambda}-f_{\mathcal{H}}\bigr).
\end{equation}
On the event where $\|(T_{\mathbf{z}}+\lambda)^{-1}LM_{\psi}\|\leq 4$,
and using $\inf \psi \ge 1/D$ which implies $\|(1-\psi)/\psi\|_{\infty}\le D$, we obtain
\begin{equation}\label{pre_ts_II_term}
\|\mathrm{II}\|_{\rhote_X}
\le
4\,\|f_{\mathcal{H}}^{\eta}-f_{\mathcal{H}}\|_{\rhote_X}
+
4\|\left(T_{\mathbf{z}}+\lambda\right)^{-1}(T-LM_{\psi})\|\,\|f_{\lambda}-f_{\mathcal{H}}\|_{\rhote_X}.
\end{equation}

Let us verify $\|(T_{\mathbf{z}}+\lambda I)^{-1}LM_{\psi}\|\le 4.$ Note, that
\begin{align*}
    \|(T_{\mathbf{z}}+\lambda I)^{-1}LM_{\psi}\| &= \|\left(I - (LM_{\psi}+\lambda)^{-\frac{1}{2}}\left(T_{\mathbf{z}} - LM_{\psi}\right)\right)^{-1}(LM_{\psi}+\lambda)^{-1}LM_{\psi}\|\\
       & \le \frac{1}{1-S''_3}, 
\end{align*}
where 
\[
S''_3:=\|(LM_{\psi}+\lambda)^{-1}(T_{\mathbf{z}} - LM_{\psi})\|.
\]
Now, applying Bernstein concentration \ref{bernstein} with random variable $\xi''_3 = (T+\lambda)^{-1} v_Y(y)\,
               K_{x}\!\left\langle K_{x},\,\cdot\,\right\rangle_{\mathcal{H}},$ we obtain, with probability as least $1-\delta/3$,
\[
S''_3 \leq \frac{2V_Y \log \left(\frac{6}{\delta}\right)}{n \lambda} + \sqrt{\frac{2\gamma_Y^2 \mathcal{N}(\lambda)\log \left(\frac{6}{\delta}\right)}{n\lambda}}.
\]
The regularisation condition $n\lambda \geq 16(V_Y+\gamma_Y^2)\mathcal{N}(\lambda)D^2\log^2(6/\delta)$,
yields $S''_3 \leq 3/4,$ therefore $\|(T_{\mathbf{z}}+\lambda I)^{-1}LM_{\psi}\| \leq \frac{1}{1-S''_3} \leq 4.$

It remains to bound $\|(T_{\mathbf{z}}+\lambda I)^{-1}(T-LM_\psi)\|$. Note that $T-LM_\psi \preceq D\,LM_\psi$, and consequently
\[
\|(T_{\mathbf{z}}+\lambda I)^{-1}(T-LM_\psi)\|
\le D\,\|(T_{\mathbf{z}}+\lambda I)^{-1}T_\psi\|
\le D\,\|(T_{\mathbf{z}}+\lambda I)^{-1}LM_\psi\|
\le 4D,
\]
where we used the bound $\|(T_{\mathbf{z}}+\lambda I)^{-1}LM_{\psi}\|\le 4$.
Substituting  to \eqref{pre_ts_II_term} yields
\begin{equation}\label{ts_II_term}
\|\mathrm{II}\|_{\rhote_X}
\le
4\,\|f_{\mathcal{H}}^{\eta}-f_{\mathcal{H}}\|_{\rhote_X}
+
4D\,\|f_{\lambda}-f_{\mathcal{H}}\|_{\rhote_X}.
\end{equation}

\paragraph{Step 4: Approximation error (Term~III).}

Proposition~\ref{approximation_error} gives directly
\begin{equation}\label{ts_III_term}
    \|\mathrm{III}\|_{\rhote_X}
    = \|f_{\lambda}-f_{\mathcal{H}}\|_{\rhote_X}
    \leq \lambda^{r}R.
\end{equation}

\paragraph{Step 5: Assembling the bound.}

Collecting \eqref{ts_I_term}, \eqref{ts_II_term}, and \eqref{ts_III_term} into \eqref{TS_excess_risk},
and using $\|f_\lambda - f_{\mathcal{H}}\|_{\rhote_X} \leq \lambda^r R$ to absorb Term~III into Term~II,
we obtain with probability at least $1-\delta$,
\begin{align*}
    \|f_{\mathbf{z},\lambda}-f_{\mathcal{H}}\|_{\rhote_X}
    &\leq
    16D(M+R)\log\!\left(\tfrac{6}{\delta}\right)
    \!\left(\frac{V_Y}{\sqrt{n\lambda}}+\gamma_Y\frac{E_s}{\sqrt{n\lambda^{s}}}\right)
    + 5D\lambda^{r}R
    + 4\left\|f_{\mathcal{H}}^{\eta}-f_{\mathcal{H}}\right\|_{\rhote_X}.
\end{align*}
With $\lambda$ chosen as in \eqref{opt_reg_imperfect_ts}, the balance condition
$n\lambda^{s} \asymp n\lambda/\lambda^{1-s}$ is satisfied and standard calculations
(see \citet{caponnetto2007optimal}) give
\[
16D(M+R)\log\!\left(\tfrac{6}{\delta}\right)
\!\left(\frac{V_Y}{\sqrt{n\lambda}}+\gamma_Y\frac{E_s}{\sqrt{n\lambda^{s}}}\right)
\leq D(M+R)\lambda^{r}.
\]
Combining and setting $C = 6D(M+R)$ yields \eqref{generalization_bound_imperfect_ts},
completing the proof.

\subsection{Proof of Theorem~\ref{IW_KRR_TarS}: Optimal IW-KRR under Target Shift}\label{proof:target_shift}

The proof follows from Theorem~\ref{main_imperfect_ts} by setting $\rho' = \rhote$. In this case, $v_Y = w_Y$, 
which implies $\phi(x) = \int_Y y d\rhote(y|x) = f_{\rhote}(x)$ and $\psi(x) = 1$. Thus $D = 1$ and the bias 
term $\|f_{\mathcal{H}}^{\eta} - f_{\mathcal{H}}\|_{\rhote_X} = 0$ vanishes.

\end{document}

%% file: math_commands.tex
\usepackage{amsmath,amsfonts,bm,amsthm}

\def\eqref#1{equation~\ref{#1}}
\def\1{\bm{1}}

\DeclareMathAlphabet{\mathsfit}{\encodingdefault}{\sfdefault}{m}{sl}
\SetMathAlphabet{\mathsfit}{bold}{\encodingdefault}{\sfdefault}{bx}{n}

\newcommand{\KL}{D_{\mathrm{KL}}}

\DeclareMathOperator*{\argmin}{arg\,min}

\newtheorem{theorem}{Theorem}
 
\newtheorem{proposition}[theorem]{Proposition} 
\newtheorem{remark}[theorem]{Remark}

\newtheorem{assumption}{Assumption}

%% file: Figures/target_shift.tex
\begin{tikzpicture}[x=0.75pt,y=0.75pt,yscale=-0.8,xscale=0.8]
%uncomment if require: \path (0,301); %set diagram left start at 0, and has height of 301

%Curve Lines [id:da20135154893253793] 
\draw    (97,201.23) .. controls (158.75,105.08) and (397.62,132.17) .. (395.99,247.27) ;
%Shape: Circle [id:dp9215261614798296] 
\draw  [dash pattern={on 0.84pt off 2.51pt}] (182.4,190.22) .. controls (182.81,162.35) and (205.74,140.08) .. (233.61,140.49) .. controls (261.49,140.9) and (283.75,163.82) .. (283.34,191.7) .. controls (282.93,219.57) and (260.01,241.84) .. (232.13,241.43) .. controls (204.26,241.02) and (181.99,218.09) .. (182.4,190.22) -- cycle ;
%Straight Lines [id:da6587143920846787] 
\draw    (215,225) ;
\draw [shift={(215,225)}, rotate = 0] [color={rgb, 255:red, 0; green, 0; blue, 0 }  ][fill={rgb, 255:red, 0; green, 0; blue, 0 }  ][line width=0.75]      (0, 0) circle [x radius= 3.35, y radius= 3.35]   ;
%Shape: Circle [id:dp9344865173416628] 
\draw  [dash pattern={on 0.84pt off 2.51pt}] (270.11,199.42) .. controls (270.4,179.11) and (287.11,162.89) .. (307.41,163.19) .. controls (327.72,163.49) and (343.94,180.19) .. (343.64,200.5) .. controls (343.34,220.8) and (326.64,237.02) .. (306.33,236.73) .. controls (286.03,236.43) and (269.81,219.73) .. (270.11,199.42) -- cycle ;
%Straight Lines [id:da14596332051278615] 
\draw    (305.87,177.96) ;
\draw [shift={(305.87,177.96)}, rotate = 0] [color={rgb, 255:red, 0; green, 0; blue, 0 }  ][fill={rgb, 255:red, 0; green, 0; blue, 0 }  ][line width=0.75]      (0, 0) circle [x radius= 3.35, y radius= 3.35]   ;
%Straight Lines [id:da5700750095528977] 
\draw    (232.87,190.96) ;
\draw [shift={(232.87,190.96)}, rotate = 0] [color={rgb, 255:red, 0; green, 0; blue, 0 }  ][fill={rgb, 255:red, 0; green, 0; blue, 0 }  ][line width=0.75]      (0, 0) circle [x radius= 3.35, y radius= 3.35]   ;
%Straight Lines [id:da043551062914208405] 
\draw    (306.87,199.96) ;
\draw [shift={(306.87,199.96)}, rotate = 0] [color={rgb, 255:red, 0; green, 0; blue, 0 }  ][fill={rgb, 255:red, 0; green, 0; blue, 0 }  ][line width=0.75]      (0, 0) circle [x radius= 3.35, y radius= 3.35]   ;

% Text Node
\draw (117.49,188.02) node [anchor=north west][inner sep=0.75pt]   [align=left] {$\displaystyle \mathcal{H}$};
% Text Node
\draw (206.75,168.01) node [anchor=north west][inner sep=0.75pt]    {$f_{\mathcal{H} }$};
% Text Node
\draw (221,213.4) node [anchor=north west][inner sep=0.75pt]    {$f_{\mathbf{z},\lambda}^{\rm IW}$};
% Text Node
\draw (309.41,166.59) node [anchor=north west][inner sep=0.75pt]    {$f_{\mathbf{z},\lambda}^v$};
% Text Node
\draw (315.75,196.01) node [anchor=north west][inner sep=0.75pt]    {$f_{\mathcal{H}}^{\eta}$};

\end{tikzpicture}